\documentclass[letterpaper, 10 pt, conference]{template/ieeeconf}  % Comment this line out if you need a4paper

\IEEEoverridecommandlockouts                              % This command is only needed if 
\usepackage{graphicx}
\usepackage{rotating}
\usepackage{adjustbox}
\usepackage{multirow}
\usepackage{color}
\usepackage{xcolor}
\usepackage{amsmath}
\usepackage{algorithm}
\usepackage{amssymb}
\usepackage[noend]{algpseudocode}
\usepackage{soul}
\usepackage{bm}
\usepackage{subcaption}
\makeatletter
\newcommand*\dotp{\mathpalette\dotp@{.5}}
\newcommand*\dotp@[2]{\mathbin{\vcenter{\hbox{\scalebox{#2}{$\m@th#1\bullet$}}}}}
\makeatother

\def\bfa{\mathbf{a}}

\def\bfg{\mathbf{g}}

\def\bfk{\mathbf{k}}

\def\bfp{\mathbf{p}}
\def\bfq{\mathbf{q}}
\def\bfr{\mathbf{r}}

\def\bfu{\mathbf{u}}
\def\bfv{\mathbf{v}}
\def\bfw{\mathbf{w}}
\def\bfx{\mathbf{x}}
\def\bfy{\mathbf{y}}
\def\bfz{\mathbf{z}}

\def\bfB{\mathbf{B}}
\def\bfC{\mathbf{C}}

\def\bfF{\mathbf{F}}
\def\bfG{\mathbf{G}}
\def\bfH{\mathbf{H}}
\def\bfI{\mathbf{I}}
\def\bfJ{\mathbf{J}}
\def\bfK{\mathbf{K}}
\def\bfL{\mathbf{L}}
\def\bfM{\mathbf{M}}

\def\bfO{\mathbf{O}}
\def\bfP{\mathbf{P}}
\def\bfQ{\mathbf{Q}}
\def\bfR{\mathbf{R}}

\def\bfT{\mathbf{T}}

\def\bfGamma{\bm{\Gamma}}

\def\bfalpha{\bm{\alpha}}

\def\bftheta{\bm{\theta}}

\def\bfsigma{\bm{\sigma}}

\def\bfomega{\bm{\omega}}

\makeatletter
\newcommand*{\itemequation}[3][]{%
  \item
  \begingroup
    \refstepcounter{equation}%
    \ifx\\#1\\%
    \else
      \label{#1}%
    \fi
    \sbox0{#2}%
    \sbox2{$\displaystyle#3\m@th$}%
    \sbox4{ \@eqnnum}%
    \dimen@=.5\dimexpr\linewidth-\wd2\relax
    \let\CenterInSpace=N%
    \ifcase
        \ifdim\wd0>\dimen@
          \z@
        \else
          \ifdim\wd4>\dimen@
            \z@
          \else
            \@ne
          \fi
        \fi
      \let\CenterInSpace=Y%
    \fi
    \ifdim\dimexpr\wd0+\wd2+\wd4\relax>\linewidth
      \@latex@warning{Equation is too large}%
    \fi
    \noindent
    \rlap{\copy0}%
    \ifx\CenterInSpace Y%
      \rlap{\hbox to \linewidth{\kern\wd0\hss\copy2\hss\kern\wd4}}%
    \else
      \rlap{\hbox to \linewidth{\hfill\copy2\hfill}}%
    \fi
    \hbox to \linewidth{\hfill\copy4}%
    \hspace{0pt}% allow linebreak
  \endgroup
  \ignorespaces
}
\makeatother

\usepackage[authormarkuptext=name,addedmarkup=bf,authormarkupposition=right]{changes}
\definechangesauthor[name={B.~L.}, color={black}]{bl}

\title{\LARGE \bf
Towards Robust State Estimation by Boosting the Maximum Correntropy Criterion Kalman Filter with Adaptive Behaviors}
\author{Seyed Fakoorian$^{*1,2}$, Angel Santamaria-Navarro$^{*2}$, Brett T. Lopez$^{*2}$,\\ Dan Simon$^1$ and Ali-akbar Agha-mohammadi$^2$% <-this % stops a space
\thanks{* Authors contributed equally to this manuscript.}
\thanks{$^{1}$Department of Electrical Engineering and Computer
Science, Cleveland State University, Cleveland, OH, USA. \texttt{\{s.fakoorian; d.j.simon\}@csuohio.edu}.}%
\thanks{$^{2}$NASA Jet Propulsion Laboratory, California Institute of Technology, Pasadena, CA, USA. \texttt{\{angel.santamaria.navarro; brett.t.lopez; aliakbar.aghamohammadi\}@jpl.nasa.gov}.}%

}

\begin{document}

\maketitle
\thispagestyle{empty}
\pagestyle{empty}

%%%%%%%%%%%%%%%%%%%%%%%%%%%%%%%%%%%%%%%%%%%%%%%%%%%%%%%%%%%%%%%%%%%%%%%%%%%%%%%%
\begin{abstract}
% Robust state estimation remains an open problem in robotics, specially when the robot is meant to navigate in perception degraded environments.
% As a step towards this robustness, in this paper we propose a new sensor fusion approach that extends the Maximum Correntropy Criterion Kalman Filter (MCCKF) by incorporating \textcolor{black}{adaptive tuning parameters, such as the noise covariance matrices or the Kernel Bandwith (KB).}
% The MCCKF outperforms traditional Kalman filters in the presence of large outliers or sensor failures. 

This work proposes a resilient and adaptive state estimation framework for robots operating in perceptually-degraded environments.
The approach, called Adaptive Maximum Correntropy Criterion Kalman Filtering (AMCCKF), is inherently robust to corrupted measurements, such as those containing jumps or general non-Gaussian noise, and is able to modify filter parameters online to improve performance.
Two separate methods are developed -- the Variational Bayesian AMCCKF (VB-AMCCKF) and Residual AMCCKF (R-AMCCKF) -- that modify the process and measurement noise models in addition to the bandwidth of the kernel function used in MCCKF based on the quality of measurements received.
The two approaches differ in computational complexity and overall performance which is experimentally analyzed.
The method is demonstrated in real experiments on both aerial and ground robots and is part of the solution used by the COSTAR team participating at the DARPA Subterranean Challenge.

\end{abstract}

%%%%%%%%%%%%%%%%%%%%%%%%%%%%%%%%%%%%%%%%%%%%%%%%%%%%%%%%%%%%%%%%%%%%%%%%%%%%%%%%
\section{Introduction}

Reliable operation of autonomous systems in diverse environments stresses the importance of accurate and robust state estimation.
\textcolor{black}{The phrase ``robust estimation" is broad (\cite{Zoubir_robust}) but can be generally categorized as algorithms that can detect and mitigate the effects of measurement dropout, divergence, or discontinuities, i.e., jumps~\cite{Santamaria-navarro2019}.}
Several modifications to the well-known Kalman filter and its nonlinear counterparts -- the extended Kalman filter (EKF) and unscented Kalman filter (UKF) -- have been proposed.
A recent method called maximum correntropy criterion Kalman filter (MCCKF), presented in~\cite{izanloo2016kalman,chen2017maximum} and further developed in~\cite{kulikova2017square,huang2017novel,wang2017maximum,kulikova2019chandrasekhar}, has proven to be effective at rejecting diverging and jumping measurements.
\textcolor{black}{Unlike the KF, which is derived using minimum mean square error (MMSE), the MCCKF uses the correntropy criterion~\cite{chen2017maximum} to capture the higher-order statistics of the measurement signal. 
The MCCKF will therefore outperform the KF when measurements contain non-Gaussian noise, i.e., the measurements are diverging or discontinuous. 
While the MCCKF is able to robustly handle non-Gaussian measurements, it relies on selecting filter parameters \textit{a priori}, which can lead to suboptimal performance if not done carefully.
This observation has lead to works that modify -- in an online fashion -- some~\cite{Akhlaghi2017,SeyedAdaptKB} or all~\cite{LiSensors2020} filter parameters.
While adapting all filter parameters is ideal, an approach has yet to be developed that does not potentially amplify the effects of outlier measurements.
% Some works have attempted to update a subset of these filter parameters~\cite{Akhlaghi2017,SeyedAdaptKB} while others~\cite{LiSensors2020} have tried to update all of them but at the expense of potentially amplifying the effects of outliers.
To address these shortcomings, this work develops a framework combining MCCKF with online filter adaptation techniques to form a unified robust and adaptive estimation pipeline that can be easily integrated into existing autonomy pipelines. 
}

Static noise models often employed in Kalman filtering is a common source of error in state estimation.
This has led to significant interest in adaptive Kalman filters (AKFs) where the noise model is learned online with real measurement data.
% The robotics and sensor fusion community has a significant interest in online learning of the noise model used in Kalman filtering.
AKFs can be classified as either correlation, covariance matching, maximum likelihood, or Bayesian methods~\cite{mohamed1999adaptive, Akhlaghi2017}.
The most common approach is the residual-based AKF (RAKF)~\cite{mohamed1999adaptive} which is a maximum likelihood technique that estimates the noise covariance matrix over a sliding-window under the assumption the noise is Gaussian.
More recent AKFs are based on Variational Bayesian techniques (VB) in which the noise covariance matrix is modeled as inverse Wishart distributions~\cite{huang2017novel18,huang2020slide}.
To the best of the authors' knowledge, an AKF method for non-Gaussian noise has yet to be developed.

\textbf{Contribution:} This paper presents a novel step towards robust state estimation using Kalman filters.
The proposed filter, called \textcolor{black}{adaptive-MCCKF (AMCCKF)}, combines the benefits of the AKF and MCCKF to achieve resilient and accurate odometry estimation of mobile robots.
% Our goal is to estimate the odometry of a robot in the presence of outliers and, to achieve it, we integrate the concepts of AKFs into an MCCKF design by proposing a new filter formulation called \textcolor{black}{adaptive-MCCKF (AMCCKF)}.
The proposed filter has the following features:
a) online adaptation of the noise covariance matrices (similar to AKF);
b) improved adaptation of the KB that \emph{will not} amplify the effects of non-Gaussian noise (unlike~\cite{LiSensors2020});
c) produce reliable estimates when the measurement or process noise is not Gaussian, e.g., contains jumps.
Two variants of the AMCCKF are derived: a VB-based AMCCKF and a residual-based AMCCKF.
This fusion approach is part of the state estimation framework developed by team CoSTAR\footnote{https://costar.jpl.nasa.gov} for the DARPA Subterranean Challenge\footnote{https://www.subtchallenge.com}, Cave Circuit~\cite{JFR-NeBula}.
We show the validity of the proposed approach by fusing real data from experiments with a custom micro-UAV platform and a Clearpath Husky\footnote{https://clearpathrobotics.com} robot both running a loosely-coupled sensor fusion architecture.

The rest of this paper is organized as follows.
% Section~\ref{sec:rev} reviews the standard KF background and two commonly used AKFs.
Section~\ref{sec:pre} reviews the MCCKF formulations.  
Section~\ref{sec:method} presents the VB-based AMCCKF and residual-based AMCCKF.
Section~\ref{sec:robot_model} describes the kinematic motion model and error state equations commonly used in mobile robot localization.
Section~\ref{sec:result} shows the experimental results.
Section~\ref{sec:conc} contains concluding remarks.

 
% \begin{equation}\label{eq.5}
% \hat{R}_k^k = \hat{\Gamma}_k^+ + HP_{k}H^\top,
% \end{equation}
% where $\hat{\Gamma}_k^+ = \frac{1}{N} \sum_{j=j_0}^{k} r_jr_j^\top$, and $r_k = y_k - \hat{x}_k$. 

% Now, let's use the same strategy to estimate $Q_k$ while assuming a known $R_k$.
% In this case, the ML equation is turned into
% \begin{equation}\label{eq.Q1}
% \sum_{j=j_0}^{k} tr\left\{ H^\top \left[\Gamma_j  - I_j I_j^\top\right]H \right\} = 0
% \end{equation}
% Then the estimated $Q_k$ is given as (see [] for the proof)
% \begin{equation}\label{eq.Q2}
% \hat{Q}_k^k = \frac{1}{N} \sum_{j=j_0}^{k} \tilde{x}_j \tilde{x}_j^\top + P_{k|k} - FP_{k|k-1}F^\top
% \end{equation}
% where $ \tilde{x}_k = \hat{x}_k - \hat{x}^-_k$. In the steady-state, $P_{k|k} = P_{k|k-1}$, and since $\tilde{x}_k = K_kI_k$, \eqref{eq.Q1} can be written as 
% \begin{equation}\label{eq.Q3}
% \hat{Q}_k^k = K\hat{\Gamma}_k K^\top
% \end{equation}

\section{Preliminaries}\label{sec:pre}
A standard assumption in Kalman filtering is the dynamics and measurements are corrupted by zero-mean Gaussian noise.
If this assumption is violated then another method is required.
The MCCKF uses higher-order statistics of the process and measurement noise to robustly eliminate the effects of non-Gaussian noise thereby preserving the Gaussian distribution of the states.
Before presenting the main results of this work, the maximum correntropy criterion and its use in state fusion is briefly summarized.

Consider the nonlinear system
\begin{equation}
    \begin{aligned}\label{eq.nonlinearsys}
         \bfx_k & = f(\bfx_{k-1}, \bfu_{k-1}) + \bfw_{k-1},\\
         \bfz_k & = h(\bfx_k) + \bfv_k,
    \end{aligned}
\end{equation}
where $\bfx_k\in\mathbb{R}^n$, $\bfu_k\in\mathbb{R}^p$, and $\bfz_k\in\mathbb{R}^m$ are the state vector, propagation inputs and observations, respectively.
The nonlinear dynamics and measurement model are $f:\mathbb{R}^n \rightarrow \mathbb{R}^n$ and $h:\mathbb{R}^n \rightarrow \mathbb{R}^m$, respectively.
The process and measurement noise are $\bfw_k$ and $\bfv_k$ and generally treated as zero-mean independent white noise processes with covariance matrices $\bfQ_k$ and $\bfR_k$, respectively.
The MCCKF equations to estimate the system in~\eqref{eq.nonlinearsys} are shown below (see~\cite{SeyedAdaptKB} for a more detailed derivation).

\noindent \textit{System model propagation (prior estimation):}
\begin{align}\label{eq.6510}
\hat{\bfx}_{k|k-1} & = f( \hat{\bfx}_{k-1|k-1}, \bfu_k),\\\label{eq.65101}
\bfP_{k|k-1} & = \bfF_k\, \bfP_{k-1|k-1}\, \bfF_k^\top + \bfQ_{k}.
\end{align}
\textit{Observation correction (posterior estimation):}
\begin{align}\label{eq.500}
  \hat{\bfx}_{k|k} &= \hat{\bfx}_{k|k-1}+{\bfK}_{k}\tilde{\bfy}_k,\\\label{eq.444}
   \bfK_{k} &= \left(\bfP_{k|k-1}^{-1}+\bfH_k^\top\bfC_{k}\bfR_{k}^{-1}\bfH_k\right)^{-1}\bfH_k^\top\bfC_{k}\bfR_k^{-1},\\\label{eq.20066}
%   \nonumber
%   \bfC_{k} & = \textem{diag}\left[\bfG_{\sigma, k}\left(\| \hat{\bfy}_{1,k} \|_{\bfR_{1,1,k}^{-1}}\right),...\,,\right.\\
%   &\hspace{3.5em} \left\bfG_{\sigma, k}\left(\|\hat{\bfy}_{m,k} \|_{R_{m,m,k}^{-1}}\right)\right],\\\label{eq.651110}
  \bfC_{k} &= \textcolor{black}{\text{diag}}\left[\bfG_{\sigma, k}\left(\| \tilde{\bfy}_{1,k} \|_{\bfR_{1,k}^{-1}}\right),..., \right. \nonumber \\
  & \hspace{1.2cm} \left. \bfG_{\sigma, k}\left(\|\tilde{\bfy}_{m,k} \|_{\bfR_{m,k}^{-1}}\right)\right],\\ 
  \bfP_{k|k} &= \left(\bfI- \bfK_k\bfH_k\right)\bfP_{k|k-1}\left(\bfI-{\bfK}_{k}\bfH_k\right)^\top \nonumber \\ 
  & \hphantom{=} +
  \bfK_{k} \bfR_{k}(\bfK_k)^\top,\label{eq.651110}
\end{align}
\noindent where $\tilde{\bfy}_k = \bfz_{k}-h( \hat{\bfx}_{k|k-1})$ and $\tilde{\bfy}_{j,k} = \bfz_{j,k}-h_j( \hat{\bfx}_{k|k-1})$, with $j \in \{1,...,m\}$; $\bfG_{\sigma, k}( \cdot )=\textrm{exp}\left(\frac{-\| \cdot \|^2}{2\sigma_k^2}\right)$, with $\sigma_k$ the size of the KB;
$\bfK_{k}\in\mathbb{R}^{n\times m}$ is the MCCKF gain;
$\bfC_{k} \in \mathbb{R}^{m\times m}$ is the correntropy gain;
$\hat{\bfx}_{k|k-1}$ is the \textit{a priori} estimate of the state $\hat{\bfx}_{k|k}$, which integrates measurements up to (and including) time $k-1$ and has covariance $\bfP_{k|k-1}$.
The \textit{a posteriori} estimate of the state $\hat{\bfx}_{k|k-1}$ is $\hat{\bfx}_{k|k}$, which is based on measurements up to (and including) time $k$ and has covariance $\bfP_{k|k}$.
Note that $|| \cdot||_A$ denotes the $A$-weighted two-norm of a vector.
The $j$-th row of the vector resulting from $h$ is $h_j$ and, similarly, $\bfR_{{j,k}}$ is the $j$-th element on the diagonal of $\bfR_{k}$.

It can be seen that when the $j$-th measurement is disturbed by a large outlier, the $j$-th element of the correntropy gain $\bfC_{k}$ in~\eqref{eq.20066} goes to zero and prevents estimator divergence.
Further, the innovation term of the $j$-th measurement can also be shown to be zero in this scenario.
In Section~\ref{sec:method} we propose two different methods for adapting the covariance noise matrices $\hat{\bfR}_k$ and $\hat{\bfQ}_k$.
We then find an adaptive method to select the KB $\sigma_k$.
For the sake of simplicity, in the following we keep only the symbol $\hat{\cdot}$ for the main variables to estimate, i.e., $\hat{\bfx}$, $\hat{\bfsigma}$, $\hat{\bfR}$ and $\hat{\bfQ}$.

\textcolor{black}{Two comments are in order.
First, note that other cost functions besides the MCC can be incorporated into the KF to reject outliers. 
For instance, Huber, Hampel or IGG (Institute of Geodesy and Geophysics)~\cite{chang2017unified}.
However, as discussed in~\cite{wang2017maximum} and \cite{chang2017unified}, MCC-based filters will outperform these filters especially when the underlying system is disrupted by impulsive noise.
Fundamentally, this is a result of large outliers being completely removed by the MCC as opposed to only being truncated by other loss functions.
And second, the MCCKF is susceptible to the same effects of linearization error and unmodeled dynamics as the EKF when outliers are not present.
However, since linearization error and excitation of unmodeled dynamics are exacerbated with measurement outliers, the MCCKF will outperform traditional EKF algorithms since the MCCKF is able to directly compensate for non-Gaussian noise. 
Additional robustness and performance improvement are also possible by adapting the process and measurement noise covariance matrices.
These points will be supported by the experiements presented in Section~\ref{sec:result}.
}

% \textcolor{black}{For nonlinear systems, we can use a nonlinear MCCKF similarly to an extended KF (EKF) and it is based on the linearization of system/measurement equations [16]. 
% Linearization errors and unmodeled dynamics can degrade the performance of the EKF (and its variants) that can be considered as a part of process/measurement noise. 
% So, the MCCKF has the same convergence issues due to the linearization errors with the EKF when the noise is Gaussian, because the correntropy gain $C_k \approx I$ results in the MCCKF reducing to standard EKF. 
% However, if the linearization errors are considered as non-Gaussian noise, it has been shown that the MCCKF performs better than EKF [16]. We will also have smaller linearization errors in the AMCCKF compared to the MCCKF, because we are estimating the noise covaraince matrices that are caused by linearization errors. 
% Moreover, in Section~III, in order to estimate the robot states, we apply the AMCCKF to estimate the error state instead of the true state under the principle of Error State KF (ESKF). The ESKF produces less linearization errors compared to the standard EKF [17]. } 

\section{Proposed Method}\label{sec:method}
\subsection{Variational Baysian based AMCCKF}
\label{sub:vb}
%A sliding-window variational adaptive Kalman filter (SWVAKF)~\cite{huang2020slide} performs better than the RAKF~\cite{mohamed1999adaptive}, but it requires more computational efforts.

A sliding-window variational adaptive Kalman filter (SWVAKF)~\cite{huang2020slide} scheme includes two steps: the forward KF (identical with the standard KF); and the backward Kalman smoother (KS). 
\textcolor{black}{The forward KF calculates state estimates based on previous estimates of noise covariance matrices. 
The backward KS provides an approximate smoothing posterior probability density function (PDF) over the sliding-window state vectors. This PDF is later combined with the posterior PDF of noise covariance matrices to make a joint posterior PDF of $p(\hat{\bfx}_{k},\hat{\bfR}_k,\hat{\bfQ}_k|\bfz_{\varpi})$. Solving this joint PDF results in an online estimate of $\bfR_k$ and $\bfQ_k$.
} In SWVAKF,  the variational Bayesian (VB) method is used to estimate online the joint posterior PDF 
%p(\bfx_{k-\varpi:k},\bfQ_k, \bfR_k|\bfz_{1:k})$
% \begin{equation}\label{eq.VB1}
% p(\hat{\bfx}_{k-\varpi:k},\bfQ_k, \bfR_k|\bfz_{1:k}) \approx q(\hat{\bfx}_{k-\varpi:k})q(\bfQ_k)q(\bfR_k)
% \end{equation}
\begin{equation}\label{eq.VB1}
p(\hat{\bfx}_{k},\hat{\bfR}_k,\hat{\bfQ}_k|\bfz_{\varpi}) \approx q(\hat{\bfx}_{k}|\bfz_{\varpi})q(\hat{\bfR}_k|\bfz_{\varpi})q(\hat{\bfQ}_k|\bfz_{\varpi}),
\end{equation}
where $q(\cdot)$ is an approximation of $p(\cdot)$ and $\bfz_{\varpi}$ is a notation simplification of $\bfz_{k-\varpi:k}$.
Since both $\hat{\bfR}_k$ and $\hat{\bfQ}_k$ are the covariance matrices representing Gaussian PDFs with zero mean, their  posterior distributions can be modeled as inverse-Wishart PDFs (IW) to guarantee conjugate inference
\begin{align}\label{eq.VB2}
q(\hat{\bfR}_k) &= \textcolor{black}{\text{IW}}(\hat{\bfR}_k |\, \hat{b}_{k|k}, \hat{\bfB}_{k|k}),\\
q(\hat{\bfQ}_k) &= \textcolor{black}{\text{IW}}(\hat{\bfQ}_k |\, \hat{t}_{k|k}, \hat{\bfT}_{k|k}),
\end{align}
where $\hat{t}_{k|k}$ and $\hat{b}_{k|k}$ are the parameters related to the \emph{degrees-of-freedom} (DoFs); and $\hat{\bfT}_{k|k}$ and $\hat{\bfB}_{k|k}$ are the inverse scale matrices,
% For the sake of simplicity, we refer the reader to~\com{[citation]} for the definitions of $\hat{t}_{k|k}$, $\hat{b}_{k|k}$, $\hat{\bfT}_{k|k}$ and $\hat{\bfB}_{k|k}$.
% In traditional SWVAKF, the estimate of $\bfQ_k$ and $\bfR_k$ can be obtained as 
% \begin{align}\label{eq.VB6}
% \bfQ_k &= \hat{\bfT}_{k|k} / \hat{t}_{k|k}\\\label{eq.VB6_1}
% \bfR_k &= \hat{\bfB}_{\bfk|k} / \hat{b}_{k|k}
% \end{align}
all described by
\begin{align}\label{eq.VB3}
\hat{t}_{k|k} &= \rho\, \hat{t}_{k|k-1} + \varpi, \quad \hat{\bfT}_{k|k} = \rho \hat{\bfT}_{k|k-1} +\sum_{j=k-\varpi}^{k} \bfO_j,\\\label{eq.VB3_1}
\hat{b}_{k|k} &= \rho\, \hat{b}_{k|k-1} + \varpi, \quad \hat{\bfB}_{k|k} = \rho \hat{\bfB}_{k|k-1} +\sum_{j=k-\varpi}^{k} \bfM_j ,
\end{align}
\noindent where $\rho$ is the forgetting factor and recommended to be in the interval $\rho \in [0.9, 1]$.
The matrices $\bfO_j$ and $\bfM_j$ are 
\begin{align}\label{eq.VB4}
\nonumber
\bfO_j &= \bfP_{j|k} -\bfF_j\bfP_{j-1,j|k}- \bfP_{j-1,j|k}^\top \bfF_j^\top \\
&\hphantom{=} + \bfF_j\bfP_{j-1,j|k}\bfF_j^\top + \tilde{\bfx}_{j|k}\tilde{\bfx}_{j|k}^\top,  \\\label{eq.VB5}
\bfM_j &= \bfr_j\bfr_j^\top + \bfH_j\bfP_{j|k}\bfH_j^\top.
\end{align}
Estimates $\hat{\bfR}_k$ and $\hat{\bfQ}_k$ can then be obtained \textcolor{black}{(\cite{huang2020slide})}
\begin{align}\label{eq.VB6}
\hat{\bfR}_k &= \hat{\bfB}_{\bfk|k} / \hat{b}_{k|k},\\
\hat{\bfQ}_k &= \hat{\bfT}_{k|k} / \hat{t}_{k|k}.\label{eq.VB6_1}
\end{align}

The SWVAKF can be combined with the MCCKF to improve its robustness to outliers.
From~\eqref{eq.VB1}, $q(\hat{\bfx}_{k})$ can be approximated as a normal Gaussian distribution, i.e., $q(\hat{\bfx}_{k}) = \mathcal{N}(\hat{\bfx}_{k}| \hat{\bfx}_{k|k}, \bfP_{k|k})$,
% \begin{equation}\label{eq.VB10}
% q(\hat{\bfx}_{k}) = N(\hat{\bfx}_{k}| \hat{\bfx}_{k|k}, \bfP_{k|k})
% \end{equation}
where the mean vector $\hat{\bfx}_{k|k}$ can be obtained under the MCC principles~\cite{chen2017maximum}. 
Thus, we can use~\eqref{eq.6510}--\eqref{eq.651110} instead of the forward KF step in the SWVAKF (see~\cite{huang2020slide} for further details on SWVAKF design).
Since the SWVAKF was derived based on the assumption that the noise has a Gaussian distribution, we propose to use the correntropy gain in~\eqref{eq.VB5} to reduce the effect of corrupted measurements while estimating $\bfR_k$, e.g., when the noise is non-Gaussian.
Defining the \emph{unweighted} correntropy gain $\bfL_k:=\text{diag}\left[\bfG_{\sigma, k}\left(\| \tilde{\bfy}_{1,k} \|\right),..., \bfG_{\sigma, k}\left(\|\tilde{\bfy}_{m,k} \|\right)\right]$, we can approximate the auxiliary matrix $\bfM$~\eqref{eq.VB5} by right and left multiplying the unweighted correntropy gain $\bfL_k$ with the first term leading to 
\begin{align}\label{eq.VB11}
\bfM_j &= \bfL_j\bfr_j\bfr_j^\top \bfL_j + \bfH_j\bfP_{j|k}\bfH_j^\top.
\end{align}
Similar to the above-derived residual-based AMCCKF, with \mbox{$\bfL_j \approx \bfI$} (e.g., in the presence of Gaussian noise)~\eqref{eq.VB11} reduces to~\eqref{eq.VB5}.
However, with non-Gaussian noise \mbox{$\bfL_k \approx \bf0$} \textcolor{black}{ which reduces  the impact of }corrupted measurements in the update of $\hat{\bfR}_k$.
Note that the backward KF step in the VB-based AMCCKF is the same as in the traditional SWVAKF.
Algorithm~\ref{tab:vb-amcckf} shows the modified VB-based AMCCKF design. 

Before concluding, it is instructive to briefly discuss the computational complexity of the SWVAKF and VB-AMCCKF.
\textcolor{black}{The computation complexity of both SWVAKF and VB-AMCCKF is highly dependant  on the window size in the KS step. The computation grows linearly as the $\varpi$ gets larger~\cite{huang2020slide}.  It is suggested to choose $\varpi$ within 5 to 20 range for optimal computation complexity~\cite{huang2020slide}.}

\begin{algorithm}[t]
\caption{Variational Bayesian-based AMCCKF}\label{tab:vb-amcckf}
\begin{algorithmic}[1]
\Statex \textbf{Initialization:}
\State $\hat{\bfx}_0, \bfP_0, \hat{\bfR}_0, \hat{\bfQ}_0,  {t}_0, {\bfT}_0, {b}_0,$ and ${\bfB}_0$ 
\Statex
\Statex \textbf{ Forward MCCKF:}  Using~\eqref{eq.6510}--\eqref{eq.651110}
\State   $[\hat{\bfx}_{k|k}, {\bfP}_{k|k}] = MCCKF (\hat{\bfx}_{k-1|k-1}, {\bfP}_{k-1|k-1},$ 
\Statex \qquad  \qquad  \qquad $\hat{\bfQ}_{k-1}, \hat{\bfR}_{k-1},  \bfF_k, \bfH_k, \bfz_k) $
\Statex
\Statex \textbf{ Backward KS:}
  \For{\texttt{j = [k-$\varpi$ : k]}}
         \State $\bfP_{j|j-1} = \bfF_k \bfP_{j-1|j-1} \bfF_k^\top + \hat{\bfQ}_j$
         \State  $\bfG_{j-1} = \bfP_{j-1|j-1} \bfF_k^\top \left(\bfP_{j|j-1} \right)^{-1}$
         \State $\hat{\bfx}_{j-1|k} = \hat{\bfx}_{j-1|j-1} + \bfG_{j-1} \left(\hat{\bfx}_{j|k} - \bfF_k\hat{\bfx}_{j|j-1} \right)$
         \State Obtain $\bfO_j$ and $\bfM_j$ using~\eqref{eq.VB4} and~\eqref{eq.VB11}
  \EndFor
\State Obtain $\bfM_{k-\varpi}$ using~\eqref{eq.VB11}
\Statex
\Statex \textbf{Estimation of $\bfQ_k$ and $\bfR_k$:}
\State Obtain the Wishart parameters using~\textcolor{black}{\eqref{eq.VB3}-\eqref{eq.VB3_1}}
\State Update $\hat{\bfR}_k$ and $\hat{\bfQ}_k$ using~\eqref{eq.VB6}-\eqref{eq.VB6_1}
\end{algorithmic}
\end{algorithm}

\subsection{Residual-based AMCCKF}

\textcolor{black}{The SWVAKF presented in Section~\ref{sub:vb} can be computationally expensive due to the backward KS step.
This section presents a residual-based approach that has lower computational complexity with similar accuracy making it more suitable for mobile robots with limited onboard computation.}
The traditional residual-based adaptive KF (RAKF) estimates $\bfR_k$ and $\bfQ_k$ assume the innovation sequence of the Kalman filter is a white process~\cite{huang2017novel18}.
The RAKF then solves a maximum likelihood (ML) problem~\cite{mohamed1999adaptive}.
We propose to modify the ML problem by incorporating the correntropy gain directly in the ML optimization.
In this case, the probability density function of the measurements $\bfz$ with the adaptive parameter $\bfalpha$ at the epoch $k$ is
\begin{equation}\label{eq.RRAKF000}
p(\bfz|\bfalpha) = \frac{1}{\sqrt{(2\pi)^m |\Gamma_k|}} e^{\frac{-1}{2}\bfL_k \bfy_k \bfGamma_k^{-1} \bfy_k^\top \bfL_k},
\end{equation}
where $m$ is the number of measurements; $\bfL_k$ is the correntropy gain~\eqref{eq.20066}, without the weighting matrix $\bfR_k^{-1}$; and $\bfGamma_k$ is the covariance matrix of the innovation sequence, which is dependent on the adaptive parameter $\alpha$. 
Since $\bfGamma_k := \mathrm{E}\left[\bfy_k\bfy_k^\top\right]$ then direct substitution gives~\cite{brown1992introduction}
\begin{align}\label{eq.RRAKF003}
 \bfGamma_k = \bfH\bfP_{k|k-1}\bfH^\top + \hat{\bfR}_k.
\end{align}

The main principle behind an adaptive Kalman filter is to determine $\bfGamma_k$ and its partial derivative with respect to $\alpha$.
One can then formulate an online stochastic natural gradient descent on the log-likelihood of the observations.
Moreover, we are interested in averaging the adaptation of $\hat{\bfR}_k$ and $\hat{\bfQ}_k$ over a sliding window of length $\varpi$.
First, we take the natural logarithm of~\eqref{eq.RRAKF000} so
\begin{equation}\label{eq.RRAKF001}
\bfJ = \sum_{j=k-\varpi}^{k} \text{ln}|\bfGamma_j| + \sum_{j=k-\varpi}^{k} \bfL_j \bfy_j \bfGamma_j^{-1} \bfy_j^\top \bfL_j.
\end{equation}
Then, maximizing~\eqref{eq.RRAKF000} translates into minimizing~\eqref{eq.RRAKF001} with respect to $\alpha$.
Therefore, one can compute the following gradient
\textcolor{black}{
\begin{equation}\label{eq.RRAKF002}
\frac{\partial \bfJ}{\partial \bfalpha_k} =  \sum_{j=j_0}^{k} \left[\text{tr} \left( \bfGamma_j^{-1} \frac{\partial \bfGamma_j}{\partial \bfalpha_k} \right) -  \bfL_j \bfy_j \bfGamma_j^{-1} \frac{\partial \bfGamma_j}{\partial \bfalpha_k} \bfGamma_j^{-1}  \bfy_j^\top \bfL_j\right].
\end{equation}
}Substituting~\eqref{eq.RRAKF003} and its derivative (see~\cite{mohamed1999adaptive}) into~\eqref{eq.RRAKF002}, the ML equation for the AMCCKF results in
\begin{align}\label{eq.RRAKF004}
\nonumber
\sum_{j=k-\varpi}^{k} \text{tr} \bigg(\left[\bfGamma_j^{-1} -  \bfL_j \bfGamma_j^{-1} \bfy_j  \bfy_j^\top \bfGamma_j^{-1} \bfL_j \right]\left[ \frac{\partial \hat{\bfR}_j}{\partial \bfalpha_k}  \right. \bigg. \\
\left. \left. + \bfH_j \frac{\partial \hat{\bfQ}_{j}}{\partial \bfalpha_k}\bfH_j^\top\right]\right)  = 0.
\end{align}
To obtain an explicit equation for $\hat{\bfR}_k$, one first assumes that $\hat{\bfQ}_k$ is known, and that $\hat{\bfR}_{k}$ has values only along its main, i.e., $\hat{\bfR}_{i,k} = \alpha_i$.
Then, \eqref{eq.RRAKF004} reduces to 
\begin{align}\label{eq.RRAKF005}
\sum_{j=k-\varpi}^{k} \text{tr} \left(\bfGamma_j^{-1}  \left[\bfGamma_j -  \bfL_j \bfy_j  \bfy_j^\top  \bfL_j \right]\bfGamma_j^{-1}\right)  = 0.
\end{align}
The solution to~\eqref{eq.RRAKF005} can be obtained under a sliding-window (average) as
\begin{align}\label{eq.RRAKF006}
\bfGamma_k  = \frac{1}{\varpi}\sum_{j=k-\varpi}^{k} \bfL_j \bfy_j  \bfy_j^\top  \bfL_j.
\end{align}
Hence, the adaptive innovation-based expression for $\hat{\bfR}_k$ can be achieved by substituting~\eqref{eq.RRAKF006} into~\eqref{eq.RRAKF003}, resulting in
\begin{align}\label{eq.RRAKF007}
 \hat{\bfR}_k = \bfGamma_k - \bfH\bfP_{k|k-1}\bfH^\top .
\end{align}

To guarantee that $\hat{\bfR}_k $ remains positive-definite, we can adapt~\eqref{eq.RRAKF006} and~\eqref{eq.RRAKF007} to use the residual (i.e., residual-based adaptive method) $\bfr_k = \bfz_k - h(\hat{\bfx}_{k|k})$, obtaining 
\begin{align}
\bfGamma_k &= \frac{1}{\varpi} \sum_{j=k-\varpi}^{k} \bfL_j\bfr_j\bfr_j^\top\bfL_j, \label{eq.RRAKF008a}\\
\hat{\bfR}_k& = \bfGamma_k + \bfH\bfP_{k|k}\bfH^\top . \label{eq.RRAKF008b}
\end{align}
The reader is referred to Appendix A for more details on the derivation of~\eqref{eq.RRAKF008a}-\eqref{eq.RRAKF008b}.

If non-Gaussian noise is present within the sliding-window, for example in one of the measurements, then the residual is no longer normal leading to an accumulative error in traditional RAKF.
As a result, the estimated $\hat{\bfR}_k$ for that particular window of measurements is not accurate even though the noise for the most recent measurement might be of Gaussian distribution.
The usage of the correntropy gain $\bfL_k$ mitigates the effects preventing the divergence of the states and $\hat{\bfR}_k$.
We should also note that $\bfL_k$ is approximately the identity matrix when the noise is Gaussian ($\bfL_k \approx \bfI$), thus the equation for adapting $\hat{\bfR}_k$ is reduced to a traditional RAKF. 

To compute an adaptive $\hat{\bfQ}_k$ we can follow a similar strategy as for $\hat{\bfR}_k$, but this time assuming $\hat{\bfR}_k$ is known and $\hat{\bfQ}_{i,k} =\alpha_i$ in~\eqref{eq.RRAKF004}.
In this case, the adaptive equation for $\bfQ_k$ can be approximated as $\hat{\bfQ}_k = \bfK_k\bfGamma_k \bfK_k^\top$, with $\bfGamma_k$ given in~\eqref{eq.RRAKF006}.
The developments to obtain $\hat{\bfQ}_{k}$ are similar but omitted for brevity (see e.g.,~\cite{mohamed1999adaptive}).

\subsection{Adaptive Kernel Bandwidth selection:}
An important parameter in MCCKF often selected through trial and error with real sensor data is the kernel bandwidth (KB).
A large KB limits the ability of the filter to reject outliers while a small KB can cause slow convergence or, in some cases, divergence of the filter~\cite{chen2017maximum}.
This work proposes to \emph{adapt} the KB online based on the quality of measurements received.
Drawing inspiration from~\cite{SeyedAdaptKB} for the dynamic selection of the KB size and motivated by the adaptation of $\hat\bfR_k$ and $\hat\bfQ_k$, we define 
\begin{equation}\label{eq.KB}
\hat{\sigma}_{\mu,k} =  \left(||\bfy_{\mu,k} \bfy_{\mu,k}^\top||_{\hat{\bfR}_{\mu,k-1}} + \bfH_\mu{\bfP}_{k|k-1}\bfH_\mu^\top \right)^{-1},
\end{equation}
where $\mu$ is the index of the particular dimension of the observation being evaluated (e.g., if we observe only 3D position, $\mu \in \{1, 2, 3\}$); and ${\bfP}_{k|k-1}$ is obtained from~\eqref{eq.65101}, using $\hat{\bfQ}_k$.
Intuitively, when a measurement is not corrupted non-Gaussian noise then $\hat{\sigma}_{\mu,k}$ in~\eqref{eq.KB} becomes large and $\bfL_{\mu,k}, \bfC_{\mu,k} \approx \bfI$.
However, if a measurement classified as an large outlier, $\hat{\sigma}_{\mu,k}$ will become small small and $\bfL_{\mu,k}, \bfC_{\mu,k} \approx \bf0$ preventing a catastrophic state correction.
Moreover, an adaptive KB has the advantage of measurement-specific adaptation, i.e., independent for each observation dimension. 

Notice how the adaptive KB in~\eqref{eq.KB} is the sum of two positive terms, therefore $\hat{\sigma}_{\mu,k}$ is always positive and well-defined.
This is not the case for~\cite{LiSensors2020}, where they proposed a KB adaptation that can reach negative values thus creating an inconsistency that the MCCKF cannot handle.
KB adaption can be used in both VB- and R-based AMCCKF.

\section{Robot State Estimation on SE(3) } \label{sec:robot_model}

A critical capability for autonomous mobile robots is robust ego-motion estimation.
To this end, we now show how the presented AMCCKF schemes can be used to robustly localize a mobile platform.
Existing filter-based localization approaches can be classified as loosely- or tightly-coupled.
The main difference being the type of measurements used in the filter.
% that in tightly-coupled architectures the sensor observations directly infer the robot state, while in the loosely-coupled methods the sensor observations are first tracked and mapped into a common domain before inferring the state in the filter.
% Both schemes have their advantages and disadvantages.
This work takes a loosely-coupled approach where the proposed AMCCKF algorithm is the main fusion engine that use an Inertial Measurement Unit (IMU) to propagate the nonlinear model and 6DoFs odometry estimates are used to correct this propagation (i.e., odometry estimations computed by one or several external front-end modules). 
In both propagation and correction steps we use the estimates $\hat{\bfQ}$, $\hat{\bfR}$ and $\hat{\sigma}_\mu$ described in the previous section.

% We may also want to augment the state vector by IMU biases and try to estimate them along with the main states of the robot. There are two approaches for the robot state estimation when KF methods are applied: loosly-coupled and tightly-coupled~\com{[citation]}. Here we apply our proposed AMCCKF into the loosly-coupled paradigm to fuse processed information from individual sensors. 

\subsection{System kinematics in discrete time}

State and covariance propagation requires a discretized version of the kinematic motion model of the robot to infer its true-state $\bfx$.
However, it is common to estimate the error-state instead of the true-state, where the error-state $\delta\hat{\bfx}$ values represent the discrepancy between the nominal-state $\hat{\bfx}$ and the true-state values $\bfx$~\cite{santamaria2018autonomous}. 
This type of propagation and update model is called Error-State Kalman Filter (ESKF) or, less frequently, Indirect Kalman Filter.
In the following developments we integrate the principles of ESKF within the AMCCKF design. 
The advantages of using the error-state kinematic model have been discussed in~\cite{madyastha2011extended}.

First, we define the nominal state kinematics in discrete-time with a first-order (Euler) integration (i.e., a Taylor series expansion, truncating the series at first grade) as
% The first step is to find the kinematic motion of the robot in order to estimate the true states of the robot. However it is common to estimate the error-states instead of true state, where the error-state $\delta x$ values represent the discrepancy between the nominal $x$ and the true-state values $x_t$~\com{[citation]}. The advantages of using error-state kinematic model compared with using regular kinematic model have been discussed in []. In the error-state-based Kalman filtering we need to integrate the nominal and the error-state equations. The nominal state kinematics in discrete-time are given as~\com{[citation]}:
\textcolor{black}{
\begin{align}
\hat{\bfp}_{k}& = \hat{\bfp}_{k-1} + \hat{\bfv}_{k-1}\Delta t,\label{es11} \\
\hat{\bfv}_{k}& = \hat{\bfv}_{k-1} + [\mathcal{\hat{R}}_{k-1}(\bfa_r-\bfa_b)  + \bfg]\Delta t , \label{es22} \\ 
\hat{\bfq}_{k} &= \hat{\bfq}_{k-1} \otimes \bfq\left\{(\bfomega_r-\bfomega_b) \Delta t)\right\},\label{es55}
\end{align}
}where $\bfp$ and $\bfq$ are the robot's 3D position and orientation (expressed as Hamiltonian quaternion with convention $[q_w, q_x, q_y, q_z]$), respectively. 
The operation $\bfq\{i\}$ represents the quaternion created from the 3D rotation vector $i$ and $\mathcal{R}$ represents a 3D rotation matrix.
The 3D linear velocity vector is $\bfv$.
The raw accelerometer and gyroscope measurements are $\bfa_r$ and $\bfomega_r$ with biases $\bfa_b$ and $\bfomega_b$.
Note that the biases are not directly estimated in this work as the IMU used in the experiments estimates them internally.
Thus the corrected measurements produced by the IMU are denoted as $\bfa_s := \bfa_r-\bfa_b$ and $\bfomega_s := \bfomega_r-\bfomega_b$.
% As IMU measurements suffer from biases, we also include in our state the accelerometer (${\bfa}_b$) and gyroscope (${\bfomega}_{b}$) biases to estimate them over time.
Finally, $\Delta t$ is the discretization time step (time differential).

The error state dynamics of the discrete-time kinematic model are given by
\begin{align}\label{eq.ES2}
\delta\hat{\bfp}_{k}& = \delta\hat{\bfp}_{k-1} + \delta\hat{\bfv}_{k-1}\Delta t,\\\label{eq.ES3}
\delta\hat{\bfv}_{k}& = \delta\hat{\bfv}_{k-1} - \left(\left[\mathcal{\hat{R}}_{k-1} \bfa_s  \right]_{\times} \delta\hat{\theta}_{k-1} \right) \Delta t,\\\label{eq.ES4}
\delta\hat{\bftheta}_{k} & = \delta\hat{\bftheta}_{k-1},
\end{align}
where we use a minimal orientation error $\delta{\bftheta} \in \mathfrak{so}(3) \subset \mathbb{R}^3$ living in the tangent space tangent of $SO(3)$ manifold (i.e., in its Lie algebra $\mathfrak{so}(3)$).

% Notice how, for the sake of clarity, we removed in~\eqref{eq.ES2}-\eqref{eq.ES4} the error terms applied to the orientation, velocity and IMU biases as they are modeled with Gaussian processes with zero mean.

% is the 3D linear velocity,  is the quaternion vector with $[q_w, q_x, q_y, q_z]$, $\bfa_s$ is the linear acceleration coming from the accelerator, $w_s$ is the angular velocity coming from the gyroscope,  $\bfb_a$, and $\bfb_a$ are accelerator and gyroscope biases, and $\Delta t$ is discretisation time.   

% To obtain the error-state kinematic equations, the error is simply added for all state variables except for the orientation and linear velocity that require some non-trivial manipulation of \eqref{es11} and \eqref{es22} []. The orientation error $\delta \bftheta$  is defined in $SO(3)$
% and can be calculated in the global frame or in the local frame. Here are the error-state kinematic equations corresponding to the nominal equations~\eqref{eq.ES1}-\eqref{es55} 
% \begin{align}\label{eq.ES2}
% \delta \bfp_{k+1}& = \delta \bfp_{k} + \delta \bfv_{k}\Delta t\\
% \delta \bfv_{k+1}& = \delta \bfv_{k} -  ([\bfOmega_k (\bfa_s −\bfb_a_{k} )]_{\times} \delta \theta + \bfOmega_k \delta \bfb_a_{k} ) \Delta t\\
% \delta \bftheta_{k+1} & = \delta_k \bftheta_{k} - \bfOmega  \delta \bfb_g_{k} \Delta t \\
% \delta \bfb_a_{k+1} & = \delta \bfb_a_{k} \\ \label{es99}
% \delta \bfb_g_{k+1} & = \delta \bfb_g_{k}
% \end{align}

\subsection{AMCCKF propagation and correction}

The discrete-time kinematic model in~\eqref{es11}--\eqref{es55} is propagated at the arrival of every new IMU measurement.
Since the dynamics are nonlinear, the dynamics are linearized to obtain the Jacobian used to propagate the state covaraince matrix.
% We construct the AMCCKF prior estimation~\eqref{eq.nonlinearsys} by propagating the nominal state $\bfx_k$ at the arrival of every new IMU measurement, $\bfu_{k-1}$ in~\eqref{eq.nonlinearsys}, based on the kinematic model in~\eqref{es11}-\eqref{es55}.
% Since $f$ in~\eqref{eq.nonlinearsys} is a nonlinear function, we need to linearize it to find the transition matrix $\bfF_k$, needed in~\eqref{eq.65101}.
% The linearized model is given in~\cite{santamaria2019}.
Note that in ESKF filters the prior estimate of the error-state is considered zero (i.e., $\delta \hat{\bfx}_{k|k-1} = \bf0$) because we assume the error mean $\delta \hat{\bfx}_{k|k-1}$ starting and remaining at zero until a new correction observation is received.
Notice how the evaluation of~\eqref{eq.65101} also includes ${\bfQ}_k$, which in our case corresponds to the estimated $\hat{\bfQ}_k$ through either the VB- or residual-based AMCCKF methods described in the previous section.

% The state error kinematic motion model~\eqref{eq.ES2}-\eqref{es99} can be written in the form $\delta \bfx = f(\delta \bfx, \bfu)$, where $\bfu$ is IMU measures. Since $f(.)$ is nonlinear we need to 
%  linearize it to find the transition matrix $\bfF_k$ as needed in~\eqref{eq.65101}. The linearized model is given in~\com{[citation]}. We should note that the prior estimation of the error-state is considered zero $\delta \hat{\bfx}_{k|k-1} = 0$  because we assume the error mean $\delta \hat{\bfx}_{k|k-1}$  starts and remains at zero until a new measurement received. 

% \subsection{AMCCKF correction}

For the correction step, the filter state is updated at every arrival of an external odometry estimate.
Hence, the observation model corresponds to a direct update to the robot position, orientation and linear velocity.
The linearized observation model $\bfH_k$ is required to update the error state and to compute the adaptive covariance of the measurement noise $\hat{\bfR}_k$.
% The formulation to obtain $\bfH_k$ is straight forward (refer to~\cite{santamaria2019} for further details on error-state observation model derivations).
Computing the Jacobian of the measurement model, we can evaluate the AMCCKF correction equations from~\eqref{eq.500}-\,-\eqref{eq.651110} with the respective adaptive matrices $\hat{\bfR}_k$.
Note that while we adapt only one $\hat{\bfQ}_k$ for the propagation step, we have to dynamically compute one $\hat{\bfR}_k$ for each observation source. 
For example, if we are accepting corrections from three external odometry estimation sources, we need to adapt three different $\hat{\bfR}_k$.
We should note that the odometry measurement sensors are independent and AMCCKF is able to fuse them based on the arrival of each sensor.
% \begin{align}\label{eq.ES3}
% \delta \hat{\bfx}_{k|k} = \bfK_k(\bfz - h(\hat{\bfx}_t))
% \end{align}

A final step of the filter after every correction is to update the nominal-state $\hat{\bfx}_k$ with the observed error state $\delta\hat{\bfx}_k$ using the composition $\hat{\bfx}_{k} = \hat{\bfx}_{k-1} \oplus \delta \hat{\bfx}_{k|k} $, where the operator $\oplus$ stands for a generic addition operation to the different state components.
Since $\delta \hat{\bfx}_{k}$ is reset to zero at the beginning of each new filter iteration, one can obtain the estimate of the true state as $\hat{\bfx}_t = \hat{\bfx}_k$.

\begin{figure}[t!]
  \centering
  \includegraphics[trim={0 0 0 0},clip, width=\columnwidth]{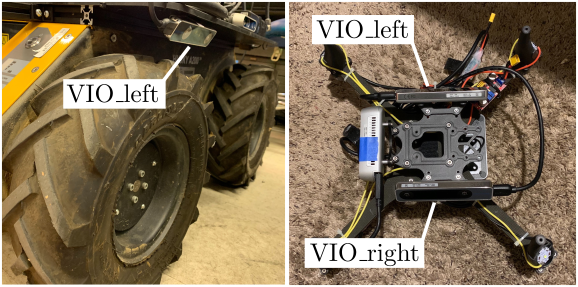}
  \caption{Clearpath Husky robot (left) and NASA-JPL small UAV platform (right and upside down) used in the experiments, both equipped with RealSense T265 camera trackers and Vectornav VN100 inertial measurements units.}
  \label{fig:robots}
  \vskip -0.1in
\end{figure}

% \begin{figure}[t]
%   \centering
%   \includegraphics[trim={0 0 0 0},clip, width=\columnwidth]{figures/pos_husky_v2.eps}
%   \begin{tabular}{c}  
%   \hspace{-0.8em}
%   \includegraphics[trim={0 0 0 0},clip,width=0.99\columnwidth]{figures/cov_husky_v2.eps}}
%   \end{tabular}
%   \caption{temporary ...}
%   \label{pos_cov_husky}
% \end{figure}

% \begin{figure}[t!]
%  \centering
%  \includegraphics[trim={0 0 0 0},clip, width=\columnwidth]{figures/pos_drone_v2.eps}
%  \caption{Position estimations during an indoor flight. We show the fusion of two RealSense-T265 camera trackers (VIO-left and VIO-right) with the proposed VB-AMCCKF. The estimation from R-AMCCKF has been avoided here for the sake of clarity as it performed like the VB-AMCCKF.}
%  \label{fig:pos_drone}
% \end{figure}

%\newpage

\begin{figure}[t!]
  \centering
  \begin{tabular}{c}
  \includegraphics[trim={0 0 0 0},clip,width=0.97\columnwidth]{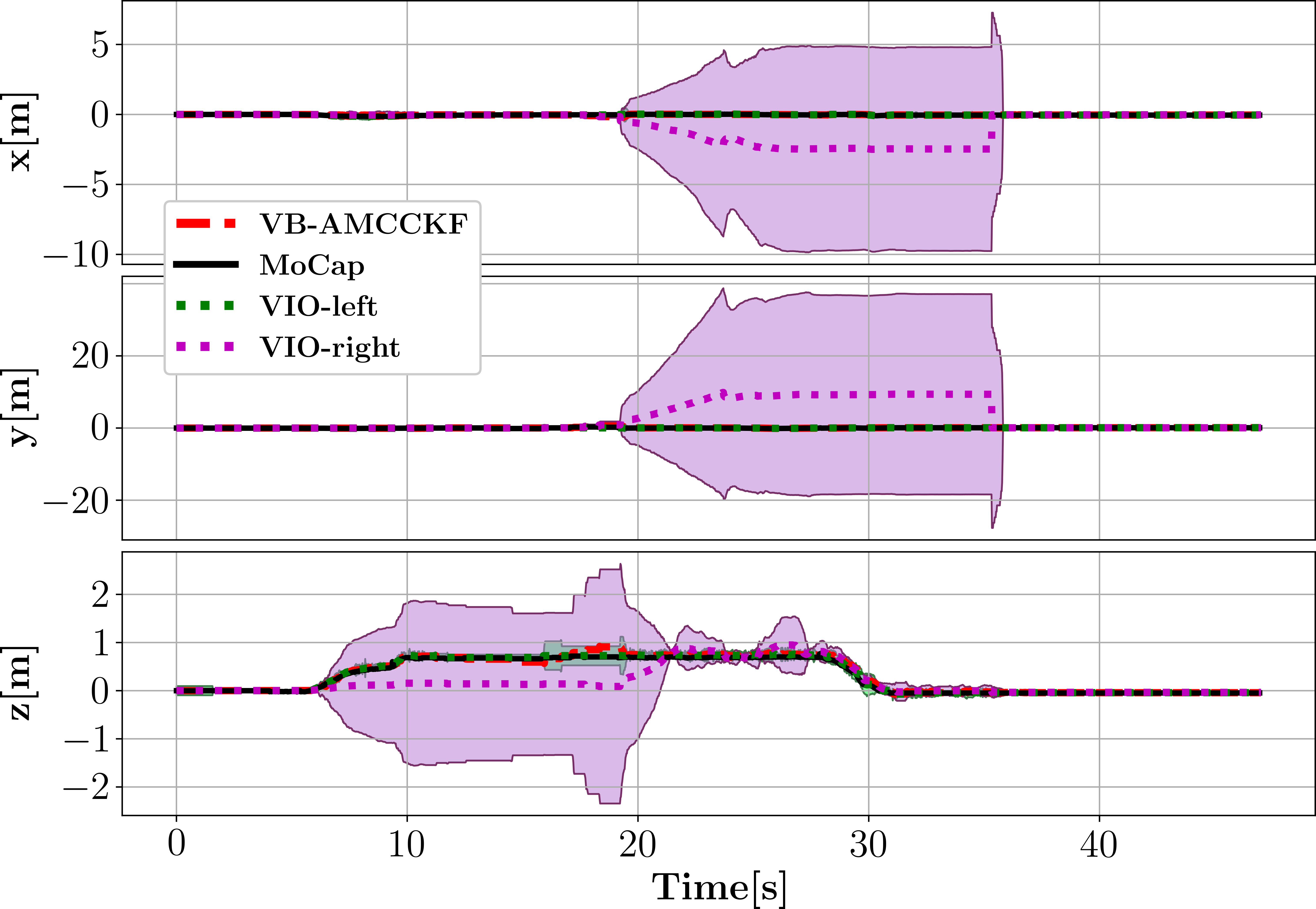}
  \end{tabular}
  \vskip -0.1in
  \caption{Position estimates from the proposed VB-AMCCKF and the estimated noise covariances for two VIO sources during an indoor flight. Example of tracking degeneracy in the right camera and how the VB-AMCCKF increases the corresponding covariance. The purple and green colored areas correspond to $\pm 3$ times the standard deviations of each tracker (i.e., \emph{3-sigma bands}), obtained from their respective $\hat{\bfR}_{i,k}$. }
  \label{fig:cov_drone}
\end{figure}

\begin{figure}[t!]
  \centering
  \begin{tabular}{l}
  \includegraphics[trim={0cm 0cm 0cm 0cm},clip,width=0.97\columnwidth]{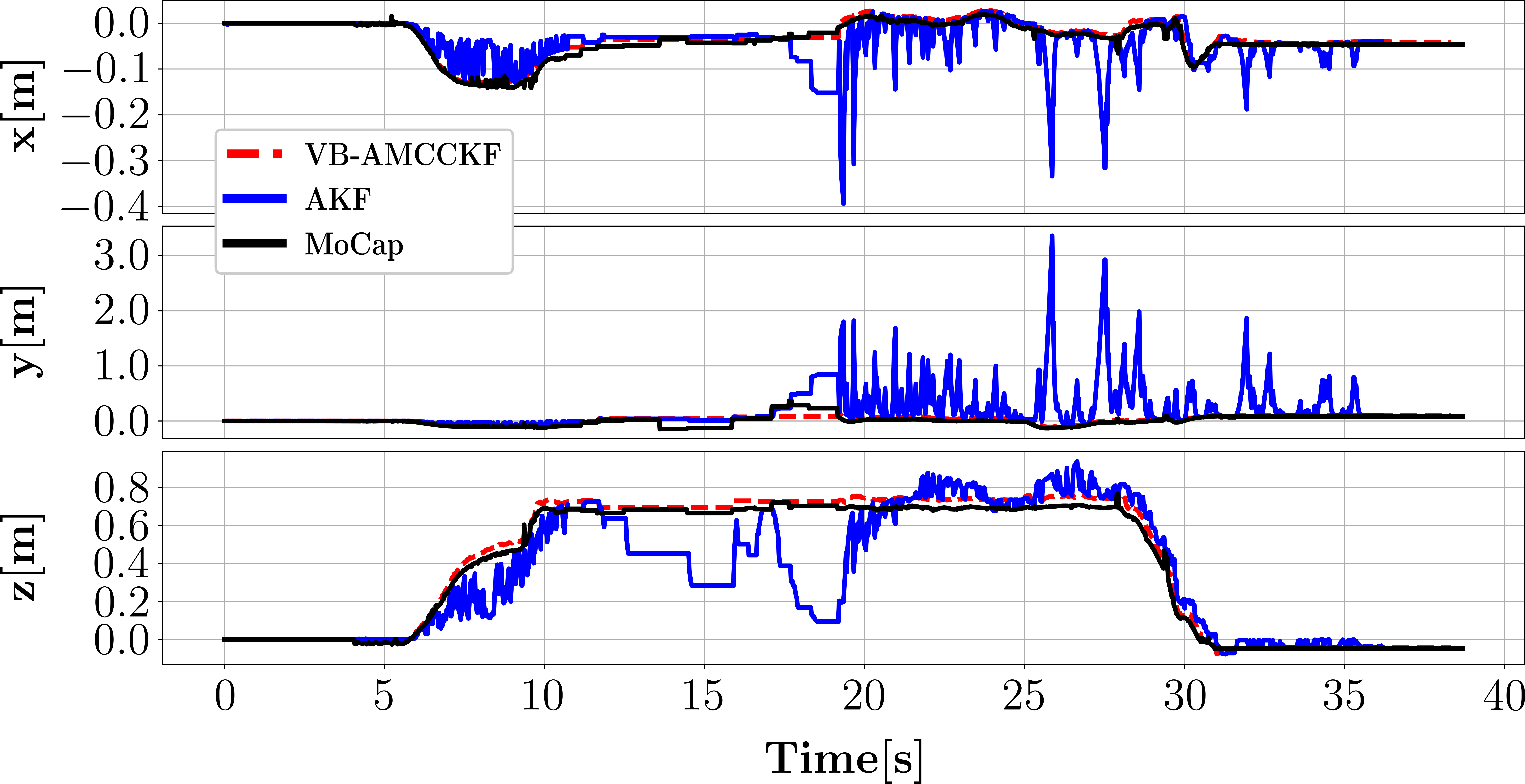}
  \end{tabular}
  \caption{Position estimates from an adaptive EKF (AKF) and the proposed VB-AMCCKF fusing two VIO sources during an indoor flight. Although the AKF has the same adaptive behavior as the VB-AMCCFK, its performance is much worse as seen by the erratic behavior of its estimate. This is attributed to the presence of non-Gaussian noise in the measurements which violates the assumptions of traditional Kalman filters.}
  \label{fig:akf_drone}
  \vskip -0.2in
\end{figure}

\begin{figure}[t!]
\begin{subfigure}{\columnwidth}
  \centering
  % include first image
  \includegraphics[trim={0 0cm 0 0},clip, width=\columnwidth]{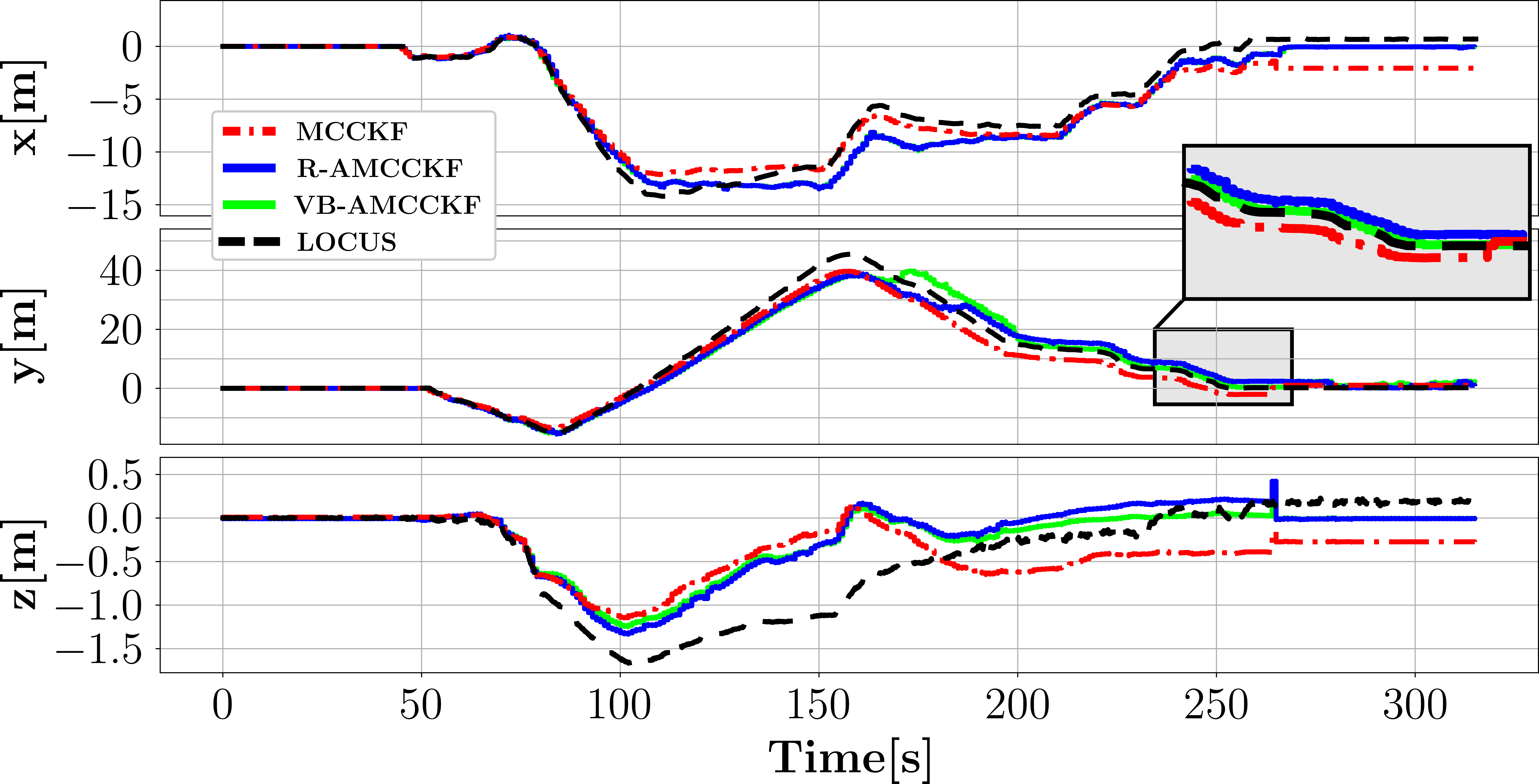}
  \vspace{-0.6cm}
  \caption{\textcolor{black}{Husky position estimations.}}
  \vspace{1em}
  \label{fig:sub-first}
\end{subfigure}
\begin{subfigure}{\columnwidth}
  \centering
  % include second image
  \begin{tabular}{c}  
  \hspace{-0.7em}
  \includegraphics[trim={0 0 0 0},clip,width=0.98\columnwidth]{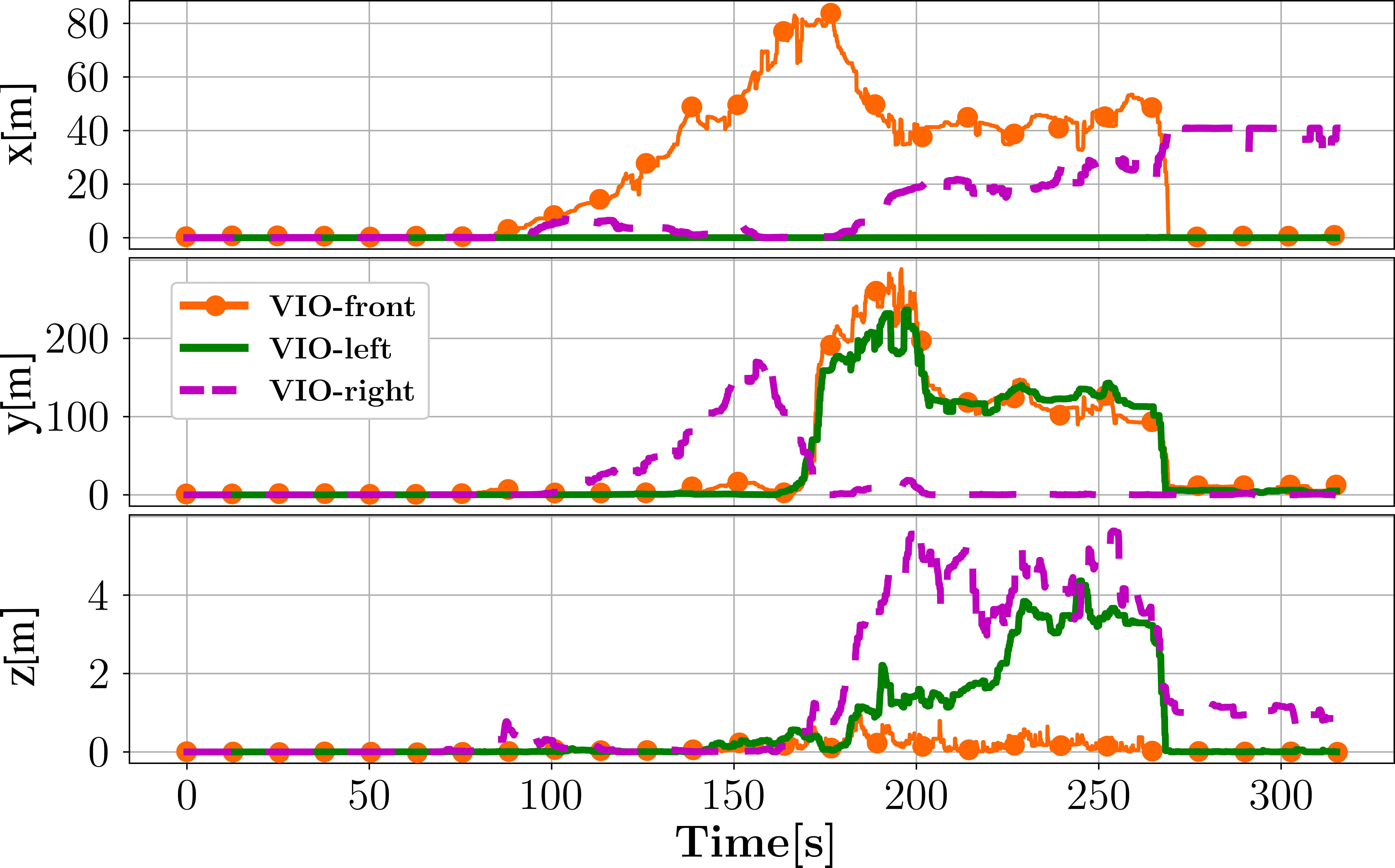}
  \end{tabular} 
  \vspace{-0.3cm}
  \caption{\textcolor{black}{Temporal variation of the adaptive kernel bandwidth (KB) inverse.}}
  \vspace{1em}
  \label{fig:sub-second}
\end{subfigure}
\begin{subfigure}{\columnwidth}
  \centering
  % include second image
  \begin{tabular}{c}  
  \hspace{-0.25cm}
  \includegraphics[trim={0 0 0 0},clip,width=0.985\columnwidth]{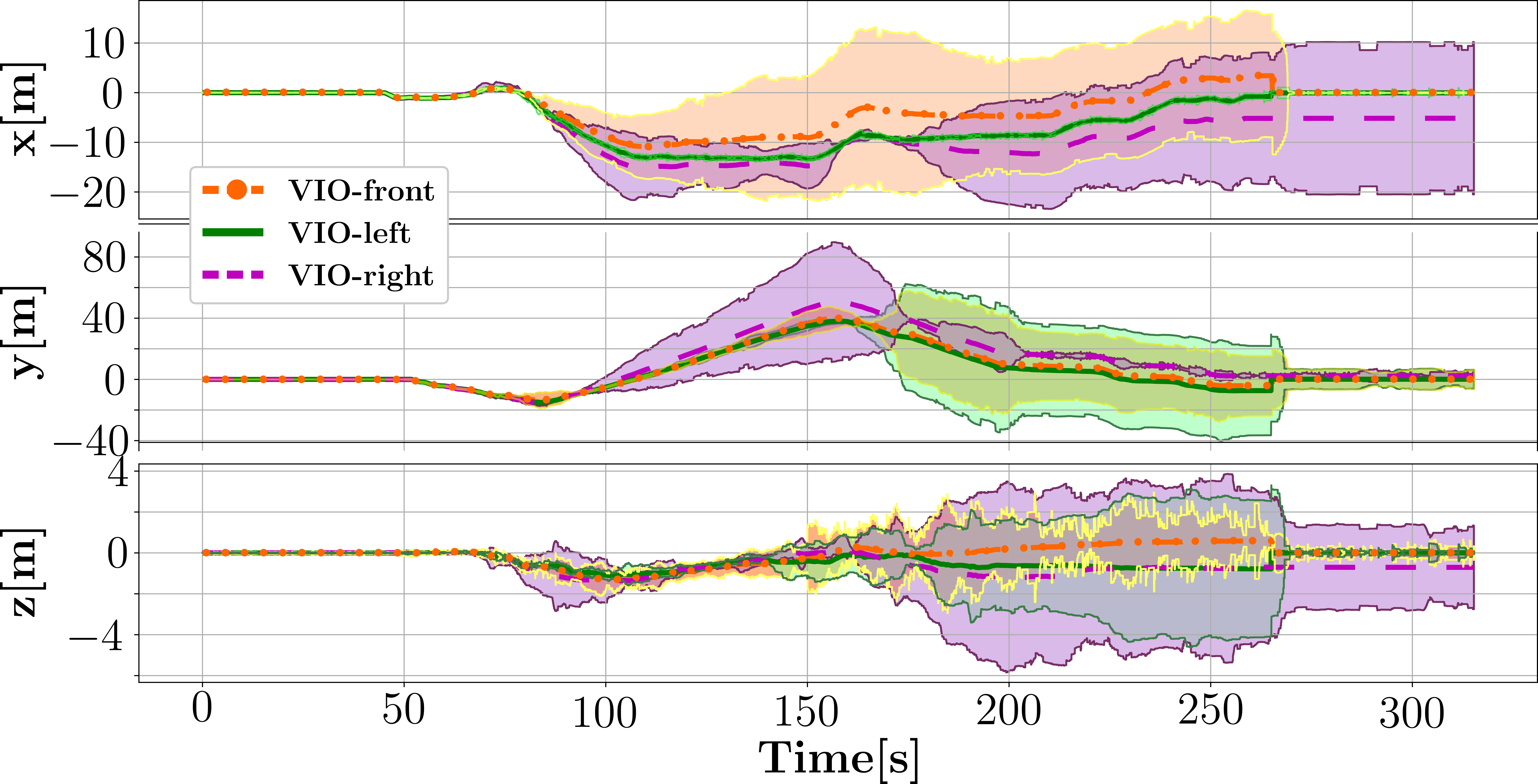}
  \end{tabular} 
  \vspace{-0.3cm}
  \caption{\textcolor{black}{The VIO estimations with their \emph{3-sigma} bands from $\hat{\bfR}_{i,k}$.}}
  %\vspace{1em}
  \label{fig:sub-third}
\end{subfigure}
  \caption{The VIO sensors are fused by VB-AMCCKF and R-AMCCKF. The position of the ground robot is shown in (a), $\pm 3$ times of the square root of estimated covariance of each VIO sensors is shown in (b), and the inverse of the adaptive KB is shown in (c). Note that we only show the position of the robot here for the purpose of illustration, however same results hold for rest of the states. }
  \label{fig:pos_cov_husky}
  \vskip -0.3in
\end{figure}

\section{Experimental Validation}\label{sec:result}

% \comb{=======================}
% \comb{The following paragraph is to have a clear idea for the text. I'll delete it once the section has been cleaned.}
% We present results highlighting the following:
% \begin{itemize}
%     \item Validity of our AMCCKF estimations by comparing with: i) Motion Capture system (MoCap) (indoor localization of an aerial vehicle); ii) state-of-the-art 3D LiDAR odometer (outdoor experiments with a Clearpath Husky ground vehicle). 
%     \item Necessity for adaptation behaviors both in KB size and in system ($\hat{\bfQ}$) and observation ($\hat{\bfR}$) covariance matrices.
%     \item The outcome from both the R-AMCCKF and VB-AMCCKF approaches.
% \end{itemize}
% \comb{=======================}

The proposed method was evaluated with real experiments with 1) a small aerial platform (UAV), designed at NASA-JPL and performing indoor flights with fast dynamics and large vibrations; and 2) a Clearpath Husky rover, able to traverse from indoor to outdoor scenarios and navigate large distances.
Both robots, shown in Fig.~\ref{fig:robots}, are part of the solution defined by the team COSTAR for the Cave Circuit in the DARPA Subterranean Challenge.
Our goal is to produce a smooth state estimate by fusing multiple odometry sources in a loosely-coupled fashion.
Commercial-of-the-shelf visual-inertial odometry (VIO) from RealSense -- the T265 camera\footnote{https://dev.intelrealsense.com/docs/tracking-camera-t265-datasheet} -- were mounted to the UAV and ground robot which can be seen in Fig.~\ref{fig:robots}.
The UAV had two T265 cameras while the Husky had three; no cameras had overlapping field-of-view so they were treated as independent measurements.
Both robots were also equipped with a Vectornav VN100 IMU sensor\footnote{https://www.vectornav.com}.
The RealSense T265 and Vectonav IMU operate at 200Hz.
Note that the gyro and accelerometer biases are not estimated since the VN100 IMU used in the experiments compensates for them internally.

In the following experiments, the kinematic model was propagated with the IMU and corrected with each VIO source.
% The AMCCKF estimates are shown in the following are obtained by propagating the kinematic model using IMU measurements and correcting the resulting states using each VIO source.
The noise covariance matrix for each camera was initialized to $\hat{\bfR}_0 = 0.01 \times \bfI_{9}$.
In addition, a window size of $\varpi = 10$ was chosen for both the VB-AMCCKF and R-AMCCKF.
Other parameters for the VB-AMCCKF were set to $\rho = 0.97$, $\hat{t}_{0|0} = \hat{b}_{0|0} = 0$, $\hat{\bfT}_{0|0} = \bf0_{9}$, and $ \hat{\bfB}_{0|0} = \bf0_{9}$.
Even though both platforms carry other sensors, only the 3D LiDAR (Velodyne VLP-16\footnote{https://velodynelidar.com/products/puck}) mounted on the Husky was used for comparison purposes.
The following experiments were carried out at the NASA Jet Propulsion Laboratory (JPL).

\subsection{UAV Experiments}

%The first set of plots (Fig.~\ref{fig:pos_drone} and~\ref{fig:cov_drone}) correspond to flights in the JPL arena equipped with a Motion Capture system\footnote{https://www.vicon.com} (MoCap).

%Specifically, Fig.~\ref{fig:pos_drone} shows the comparisons of the position estimations from the MoCap, the left and right camera trackers (VIO-T265) and the VB-AMCCKF approach proposed in this paper.
%This plot shows how, under nominal circumstances (i.e., both camera trackers are performing with the expected accuracy), the output of the AMCCKF produces reasonable filtering estimates by averaging the outputs of both camera trackers.
%However, these nominal circumstances are quite fragile in real world scenarios were situations with perception degradation are common.

Fig.~\ref{fig:cov_drone} compares position estimates during a test flight of the UAV from the left and right camera trackers (T265 camera), the VB-AMCCKF approach proposed in this paper, and a Motion Capture (MoCap) system\footnote{https://www.vicon.com}.
\textcolor{black}{The  robot  mainly  hovered  with  a few other maneuvers. 
% Robustness to poor measurements and adapting filter parameters are the two defining traits of the proposed 
% The key properties of the proposed AMCCKF methods is the ability to adapt and robustly handle 
% sensor measurements, i.e., the T265 cameras, that are under performing or failing.
In Fig.~\ref{fig:cov_drone} it is seen that the right camera started to diverge at certain points (possibly cause by poor calibration or few features in the environment).
The noise covariance  for the right camera (purple area showing \mbox{$\pm 3\sqrt{\hat{R}_{i,k}}$}, with \mbox{$i\in\{1,2,3\}$}) was subsequently increased by the VB-AMCCKF showcasing its ability to identify and adapt to failing sensors.
To determine whether only an adaptation behavior is sufficient for these types of scenarios, i.e., removing the MCC component, the VB-AMMCKF was compared to a traditional AKF.
Fig.~\ref{fig:akf_drone} shows that the AKF performs substantially worse than the VB-MCCKF as seen by noisy estimate it generates.
It is important to also note that if the AKF estimate were to be used in feedback, the system would become unstable with high probability further highlighting the advantages of the developed approach.
An EKF using the same initial noise covariance matrices as in VB-AMCCKF was also evaluated and generated estimates with several meters of error; the results are omitted for plotting clarity.
Note, the R-AMCCKF had similar performance to VB-AMCCKF and is omitted for clarity of plotting.
The results in Fig.~\ref{fig:cov_drone} and Fig.~\ref{fig:akf_drone} showcase the effectiveness of the proposed method to generate smooth and accurate estimates despite the presence of failing or corrupted measurements.
Moreover, the combination of robustness -- through MCC -- and adaptation are the keys to generating robust estimates. 
% This behavior prevented the divergence of the AMCCKF by disregarding the corrections from this sensor since increasing the noise covariance matrix (increase in shaded area) means a smaller contribution in the fusion of the signals.
% This is confirmed by 
}

% \textcolor{black}{
% In order to quantify the benefits of utilizing the MCC in Kalman filtering, an adaptive EKF (AKF) was compared to the proposed VB-AMCCKF during the same flight test.
% Fig.~\ref{fig:akf_drone} shows that the AKF performs substantially worse than the VB-MCCKF as seen by noisy estimate it generates.
% It is important to also note that if the AKF estimate were to be used in the feedback, the system would become with high probability further highlighting the advantages of the developed approach.
% The results in Fig.~\ref{fig:cov_drone} and Fig.~\ref{fig:akf_drone} showcases the effectiveness of the proposed method to generate smooth and accurate estimates despite the presence of failing or corrupted measurements.
% The R-AMCCKF had similar performance to VB-AMCCKF and is omitted for clarity of plotting.
% }

\subsection{Husky Experiments}

% For the miro-UAV the left and right VIO sensors are fused by the VB-AMCCKF. The robot was texted in the presence of the motion capture system (MoCap)\footnote{https://www.vicon.com/} in order to generate a ground truth data. The  robot  mainly  hovered  with  a few maneuvers. The VB-AMCCKF estimates are compared with the MoCap and each individual VIO sensors, as shown in Fig.~\ref{pos_cov_drone}. We can see the VB-AMCCKF is mainly the average between the two VIO sensors. However, in another test in Fig.~\ref{pos_cov_drone}, the right VIO sensor diverges after 8 sec; however, the AMCCKF prevents the divergence of the state estimates through rejecting the right VIO. 

Both the VB-AMCCKF and R-AMCCKF methods were also tested on the Husky robot during an experiment where the robot executed a several-meter traverse in an urban environment.
In this experiment ground truth from MoCap was not available so the results are compared to a LiDAR-based odometry estimation algorithm called  LOCUS~\cite{Palieri}.
\textcolor{black}{The robot was driven over non-flat terrain to intentionally degrade the performance of the T265 cameras.}

Fig.~\ref{fig:sub-first} first compares the traditional \mbox{MCCKF} (manually tuned with a fixed kernel size and covariance matrices) and the proposed VB-AMCCKF and R-AMCCKF with adaptive KB; the noise covariance matrices were kept static in this test.
Here, the traditional MCCKF cannot reach the accuracy of the VB-AMCCKF or R-AMCCKF because the MCCKF lacks adaptive behavior; both VB-AMCCKF and R-AMCCKF generate similar estimates to LOCUS at the end of the trajectory.
The inverse of the adaptive KB (Fig.~\ref{fig:sub-second}) shows the temporal weighting of the sensors in the filter.
Smaller values correspond to larger a correntropy gain (i.e, $\bfC_k \rightarrow I$ and more impact in the estimation) and vice versa.
Hence, the inverse can be interpreted as a metric for anticipated sensor health; a smaller value corresponds to a sensor that is unlikely to produce an unhealthy measurement, i.e., a healthy sensor.
VB-AMCCKF has shown better precision than R-AMCCKF but requiring more computation. 
\textcolor{black}{We compare the average percentage of the CPU usage between the VB-AMCCKF and R-AMCCKF in Table~\ref{cpu_info}. 
As expected the VB-AMCCKF requires more CPU effort.}

As previously stated, one of the advantages of the AMCCKF scheme is the adaptation of the observation noise covariances $\hat{\bfR}$.
Fig.~\ref{fig:sub-third} shows how the R-AMCCKF adapts for each VIO sensor.
The modification of individual dimensions on the covariance matrices allow the AMCCKF to take 
advantage of the ``best" available estimate for each individual dimension.
For instance, notice how VIO-left performs best in the estimation for $x$ position, while a combination of front and right cameras prevail for $y$, and VIO-front for $z$.
Fig.~\ref{fig:pos_cov_husky} further demonstrated the effectiveness of the developed VB-AMCCKF and R-AMCCKF.

\begin{table}[t]
\textcolor{black}{
  \caption{\textcolor{black}{CPU Usage Comparison.}} \label{cpu_info}
  \centering
  \begin{tabular}{|c|c||c|c|c|c|}
        % \caption{\textcolor{black}{ Comparison of the CPU usage between R-AMCCKF and VB-AMCCKF, on the computer with Inter(R) Core(TM) i7-9750H CPU @ 2.60GHz.} }\label{cpu_info}
     \hline
  & & \textbf{Min$\%$}& \textbf{Max$\%$}  & \textbf{Avg$\%$} & \textbf{Std$\%$}\\
     \hline
     \hline
        \multirow{2}{*}{\begin{turn}{} \textbf{UAV}\end{turn}}& \textbf{R-AMCCKF }& 30  & 69 & 40 & 12\\ \cline{2-6}
   & \textbf{VB-AMCCKF }& 38  & 81 & 58 & 12\\ \cline{2-6}
     \hline
     \hline
      \multirow{2}{*}{\begin{turn}{} \textbf{Husky}\end{turn}}& \textbf{R-AMCCKF }& 32  & 74 & 46 & 14\\ \cline{2-6}
   & \textbf{VB-AMCCKF }& 39  & 86 & 65 & 16\\ \cline{2-6}
   \hline
   \end{tabular}
}
%   \caption{\textcolor{black}{ CPU Usage Comparison Between R-AMCCKF and VB-AMCCKF.} }\label{cpu_info}
\vskip -0.1in
\end{table}

\section{Conclusion}\label{sec:conc}
This paper proposes two AMCCKF algorithms, based on the Variational Bayesian (VB-AMCCKF) and Residual (R-AMCCKF) techniques, as a step toward robust sensor fusion.
The two designs can be selected based on desired accuracy and available onboard computation.
% These two designs have complementary advantages, being VB-AMCCKF the more accurate but with R-AMCCKF the less computationally expensive.
Fundamentally, AMCCKF leverage the advantages of MCCKF filters, which show improved robustness with respect to traditional Kalman filters, and extended the design to incorporate adaptive behaviors.
The adaptations include a dynamic selection of a) the kernel bandwidth size; b) the system noise covariance matrix; and c) the measurement noise covariance matrix.
The addition of online adaptation reduce, the effects of rigid (and usually) inaccurate user-defined parameters.
The validity of the proposed filters was demonstrated through real experiments on a UAV and Husky ground robot.
%
% We proposed adaptive maximum correntropy Kalman filter (AMCCKF) in order to robust sensor fusion in robot odomtery estimation. The AMCCKF reduces the effect of inaccurate user-defined parameters, (e.g., noise covaraince matrices, and KB), while also reject the outlier or shot noise from the estimation pipeline, both states and covariance matrices. The VB-AMCCKF performs better than R-AMCCKF, but requires more computation because of the backward smoothing step. We showed the validity of the proposed filter through real experiments on the ground flying robots that designed and sponsored at NASA-JPL. 
%
% In these experiments we can see clear advantages of the adaptive behaviors.
% For example, when there is some redundancy in the filter inputs so the AMCCKF classifies their quality by adapting the respective noise covariances.
% Although the results with adaptive KB selection are promising, with the filters reaching better values than user-defined fixed parameters, the presented method does not guarantee the optimal value of KB. 
% Hence, we consider as future work the formulation of the KB adaptation that leads to its optimal values.

It is worth mentioning some critical points that can drive the AMCCKF to wrong estimations.
First, even though the adaptive behaviors for both system and observation noise covariances present clear benefits, there is still some dependency on their initial values which are user-defined.
These values, however, may play a role during the initial phase of the experiment, hence when the events are most likely to be under the control of the user (e.g., using a take-off platform or a docking station).
Second, the window size ($\varpi$) affects the dynamics of the filter convergence due to the \emph{averaging effect}, the larger $\varpi$ the slower the convergence.
Thus, $\varpi$ should be roughly set according to the vehicle dynamics and the purpose of the estimation.
And third, the forgetting factor $\rho$ and the values of the Wishart distribution play a critical role in the performance of the VB-AMCCKF and need to carefully selected (see e.g.,~\cite{huang2017novel18}.
Future work will pursue thorough proofs on the filter stability for both R-AMCCKF and VB-AMCCKF using Lyapunov or contraction-based stability theory.

% \com{We should also mention some critical points that may cause the AMCCKF fails. $i.$ The initial values that we assign for both covaraince of the noise terms and states are important, that may filter puts more weights on a measurement which results in the large estimation error. Of course this fact is mainly common for most Kalman-type filters. $ii.$ The window size also plays main role for the adaptive-based KFs. The large or smaller window size results in the slower convergence or increasing estimation error. Moreover, in the case of VB-AMCCKF, the initial values of Wishart distribution and the value of forgetting factor $\rho$ are important for the performance of the filter as discussed in~\cite{huang2017novel18}}

\section*{Acknowledgements}
%The authors would like to thank dr. Brett T. Lopez (NASA-JPL, California Institute of Technology) for his support on the experiments using the aerial vehicle.
This research work was carried out at the Jet Propulsion Laboratory, California Institute of Technology, under a contract with the National Aeronautics and Space Administration.
Copyright 2020 California Institute of Technology. U.S. Government sponsorship acknowledged.

\section*{Appendix A}\label{appendixA}

\noindent\textit{Residual-based AMCCKF Derivation:} From Kalman filtering theory, we can obtain

\begin{equation}\label{eq.A1}
{\bfGamma}_k^{-1} \bfy_k = \hat{\bfR}_k^{-1} \bfr_k.
\end{equation}
Substituting~\eqref{eq.A1} into~\eqref{eq.RRAKF005} and using matrix trace properties, results in
\begin{equation}\label{eq.A*}
\sum_{j=j_0}^{k} \text{tr}\left\{ \hat{\bfR}_j^{-1} \left[ \hat{\bfR}_j\bfGamma_j^{-1}\hat{\bfR}_j - \bfL_j\bfr_j \bfr_j^\top\bfL_j\right]  \hat{\bfR}_j^{-1} \right\} = 0.
\end{equation}
Further we also know (\cite{kulikova2017square}) that
\begin{equation}\label{eq.A2}
\bfP_{k-1|k}\bfH^\top \bfGamma^{-1} = \bfP_{k|k}\bfH^\top\hat{\bfR}_k^{-1}.
\end{equation}
Left-multiplying both sides of~\eqref{eq.A2} by $\bfH$ and knowing that the covariance matrix of the  innovation is \mbox{$\bfGamma_k = \bfH\bfP_{k|k-1}\bfH^\top + \bfR_k$}, we can obtain 
\begin{equation}\label{eq.A3}
\left(\bfGamma_k - \hat{\bfR}_k \right)\bfGamma_k^{-1}= \bfH\bfP_{k|k}\bfH^\top\bf\hat{\bfR}_k^{-1}.
\end{equation}
Then, right-multiplying both sides of~\eqref{eq.A3} by $\hat{\bfR}_k$ and substituting into~\eqref{eq.A*}, we have
\begin{equation}\label{eq.A4}
\sum_{j=j_0}^{k} \text{tr}\left\{ \hat{\bfR}_j^{-1} \left[ \hat{\bfR}_j-\bfH\bfP_j\bfH^T - \bfL_j\bfr_j\bfr_j\top\bfL_j\right]  \hat{\bfR}_j^{-1} \right\} = 0.
\end{equation}
The solution of~\eqref{eq.A4} results in the residual-based adaptive estimation of $\bfR_k$ 
\begin{align}
\bfGamma_k &= \frac{1}{\varpi} \sum_{j=k-\varpi}^{k} \bfL_j\bfr_j\bfr_j^\top\bfL_j, \tag{\ref{eq.RRAKF008a}}\\
\hat{\bfR}_k& = \bfGamma_k + \bfH\bfP_{k|k}\bfH^\top.\tag{\ref{eq.RRAKF008b}}
\end{align}

\bibliographystyle{template/IEEEtran}
\bibliography{references}

% \com{To following appendices will go to an external document}

\end{document}